\documentclass[twoside,leqno,twocolumn]{article}

\usepackage[letterpaper]{geometry}
\usepackage{comment} 
\usepackage{ltexpprt}
\usepackage{hyperref}
\usepackage{graphicx}
\usepackage{amsmath}
\usepackage{bm}
\usepackage{bbm}
\usepackage{amssymb}
\usepackage{enumitem}
\usepackage{wrapfig}
\usepackage{textcomp}
\usepackage{booktabs}
\usepackage{hyperref}
\usepackage{makecell}
\usepackage{multirow}

\begin{document}

\newcommand\relatedversion{}
\renewcommand\relatedversion{\thanks{The full version of the paper can be accessed at \protect\url{https://arxiv.org/abs/1902.09310}}} 

\title{\Large Label Distribution Learning-Enhanced Dual-KNN for Text Classification}
\author {
Bo Yuan \thanks{Zhejiang University. {byuan, yulinchen, wangjy77, yinzh}@zju.edu.cn}
\and Yulin Chen\footnotemark[1]
\and Zhen Tan\thanks{Arizona State University. {ztan36, huanliu}@asu.edu}
\and Wang Jinyan\footnotemark[1]
\and Huan Liu\footnotemark[2]
\and Yin Zhang\footnotemark[1] \thanks{Corresponding Author}
}

\date{}

\maketitle


\fancyfoot[R]{\scriptsize{Copyright \textcopyright\ 2024 by SIAM\\
Unauthorized reproduction of this article is prohibited}}





\begin{abstract} \small\baselineskip=9pt 
Many text classification methods usually introduce external information (e.g., label descriptions and knowledge bases) to improve the classification performance. Compared to external information, some internal information generated by the model itself during training, like text embeddings and predicted label probability distributions, are exploited poorly when predicting the outcomes of some texts. In this paper, we focus on leveraging this internal information, proposing a dual $k$ nearest neighbor (D$k$NN) framework with two $k$NN modules, to retrieve several neighbors from the training set and augment the distribution of labels. For the $k$NN module, it is easily confused and may cause incorrect predictions when retrieving some nearest neighbors from noisy datasets (datasets with labeling errors) or similar datasets (datasets with similar labels). 
To address this issue, we also introduce a label distribution learning module that can learn label similarity, and generate a better label distribution to help models distinguish texts more effectively. This module eases model overfitting and improves final classification performance, hence enhancing the quality of the retrieved neighbors by $k$NN modules during inference. Extensive experiments on the benchmark datasets verify the effectiveness of our method. 

\end{abstract}

\section{Introduction.}
Text classification is a fundamental task \cite{tan2023interpreting, DBLP:journals/kbs/ChenYLG23} in NLP (natural language processing), in which a given text is classified into one or more of $n$ categories. 
Text classification can be used for a variety of purposes, including relationship extraction \cite{DBLP:journals/corr/abs-2107-03751}, tag recommendation \cite{DBLP:journals/ijon/LeiFYL20}, and so on. 
Generally speaking, most text classification methods mainly consist of three steps: 
firstly feed the original text into a classification model to obtain text embedding; then feed the text embedding into a linear layer to obtain label prediction probability; finally, calculate loss with the distribution of predictions and real labels. 
\begin{figure}[ht!]
    \centering
    \includegraphics[width=7.8cm]{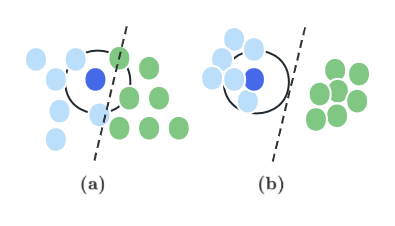}
    \caption{
    We use dots of different colors to denote different classes.
    (a) Previous work based on $k$NN may retrieve neighbors (the dots in the black circle) belonging to other classes (green dots) when the target text (deep blue dots) in similar datasets is relatively hard to distinguish (classification boundaries are close).
    (b) Our proposed label distribution learning can improve the performance of models (different class clusters get tighter and away from classification boundaries), thus the quality of retrieved neighbors in $k$NN is enhanced (the dots in the black circle are all blue).}
    \label{figure1}
\vspace{-0.3cm}
\end{figure}
However, such a standard process has its problems during inference: 
(1) The predicted results depend on the linear layer added directly on top of the text encoder, which may face the overfitting problem as pointed out by \cite{DBLP:journals/corr/abs-2110-05409}. 
(2) Only the trained parameters are used, and other forms of information that can be directly obtained from historical training instances are ignored. To tackle these problems, it is intuitive to save more historical information during training, then make predictions based on them and not solely consider the ultimate output probability of the model.


Recently, retrieval-augmented methods have attracted lots of attention in the natural language processing field, such as language modeling 
\cite{DBLP:journals/corr/abs-1911-00172}, named entity recognition \cite{DBLP:journals/corr/abs-2203-17103}, multi-label text classification \cite{wang2022contrastive}. 
However, these methods retrieve neighboring instances based on text embedding or token embedding, and do not fully consider other information in the intermediate layers of the model. Motivated by these works, we add the $k$ nearest neighbor mechanism based on the text embedding representations and predicted label probability representations to make predictions, instead of only using the ultimate output probability to make predictions.


Specifically, we propose a dual $k$NN (D$k$NN) framework for text classification tasks, which first constructs two representation stores to explicitly memorize the output representations of the model's middle layers during training and then retrieves $k$ nearest neighbors from these cached representations during inference. 
When retrieving, 
we apply two $k$ nearest neighbor modules, namely text-$k$NN and pro-$k$NN respectively. 
Intuitively, the retrieved neighbors may include noise (belongs to other classes) when the target text is relatively hard to distinguish (e.g., the texts with similar labels in the representation store have semantic similarity, or the relevant context is not enough in the representation store). 
Furthermore, we also find that the $k$NN mechanism is very sensitive to noise. Intuitively, as in the left part of Figure \ref{figure1}, the wrong retrieved results will make final prediction results poor robustness and generalization.

To address this problem, we propose to learn the label distribution based on label similarity and utilize contrastive learning to make the label distribution representation more distinctive. By doing this, both the quality of the retrieved neighbors and the classification performance of the original model are enhanced. 
We conducted extensive experiments to evaluate the effectiveness and robustness of our method. Experimental results show that our method can: (1) improve the classification performance of baseline models consistently on five benchmark datasets; (2) improve the defense ability against noise attacks significantly. Our contributions are summarized as follows:
\begin{itemize}
    \item We propose a dual $k$NN (D$k$NN) framework for text classification tasks that utilize the representations (text embedding and predicted label probability) from training datasets in the inference stage.

    \item We introduce a label distribution learning module as an effective enhancement component for D$k$NN in the retrieval process. In addition, this module can also improve the performance of base classification models. 

    \item Extensive experiments show that our proposed method produces significant improvements in baseline models. Furthermore, we conduct additional experiments to confirm the robustness of our method against noise attacks.
\end{itemize}

\section{Related Work.}
\subsection{Text Classification.}
There have been many recent works that begin  to exploit external information in text classification tasks, not only focusing on learning text representation. 
\cite{DBLP:conf/aaai/ChenHLXJ19} treats conceptual knowledge as a type of knowledge and retrieves knowledge from an external knowledge source to enhance the semantic representation of short texts. 
\cite{DBLP:conf/aaai/LiL0021} utilizes the statistical features of the corpus, such as word frequency and distribution over labels, to improve classification performance. 
\cite{DBLP:conf/aaai/ArousDYBCC21} incorporates human rationales into attention-based text classification models to improve the explainability of classification results. 
Despite all existing works, few studies pay attention to the internal information generated by the model during training. These data are based on referable historical instances, which can also aid the model in making predictions.

\subsection{Nearest Neighbor Methods in NLP.}
The $k$NN ($k$ nearest neighbors) classifier is a common basic machine learning method \cite{DBLP:conf/aaai/ZhaoL21}, and its basic idea is to determine the category of samples according to the surrounding limited adjacent samples (i.e., $k$ nearest neighbors) \cite{DBLP:journals/kbs/WangPD22}. 
Recent success on various NLP tasks has shown the effectiveness of $k$NN mechanism in improving the quality of NLP models. 
\cite{DBLP:conf/iclr/KhandelwalLJZL20} extends a pre-trained neural language model by linearly interpolating with $k$NN mechanism.
Based on $k$NN-MT proposed by \cite{DBLP:journals/corr/abs-2010-00710}, \cite{zheng2021adaptive} introduces adaptive $k$NN-MT to determine the choice of $k$ regarding each target token dynamically. 
However, we found that the $k$NN algorithm will retrieve some incorrect noise results on some datasets. 
\cite{DBLP:journals/corr/abs-2110-02523} 
also realizes that the $k$NN method tends to introduce noise in simple applications, and they both introduce contrastive learning based on text representations to construct positive and negative samples. However, our method only focuses on label representations, which are different from them. 
These previous works utilize $k$NN algorithm based on the text embedding or the contextualized word embedding. In our paper, we will introduce another novelty $k$ nearest neighbor ($k$NN) module based on the predicted label probability to consider more information generated by models during training.

\subsection{Label Distribution Learning.}
Label Distribution Learning (LDL) \cite{DBLP:journals/tkde/Geng16} is a novel machine learning paradigm for applications where the overall distribution of labels matters. A label distribution covers
a certain number of labels, representing the degree to which
each label describes the instance. \cite{DBLP:journals/tkde/Geng16} also suggests LDL could be helpful for such scenarios: some labels are correlated with other labels. Based on it, \cite{DBLP:conf/cvpr/KangJS06} 
utilizes LDL to improve the performance of multi-label text classification tasks. However, for single-label text classification tasks where they only have a unique label for each sample, the true label distribution is hard to obtain. Thus, we explore utilizing label similarity to learn label distribution, which can make models more generalized.

\begin{figure*}[t]
\centering
\includegraphics[width=16cm]{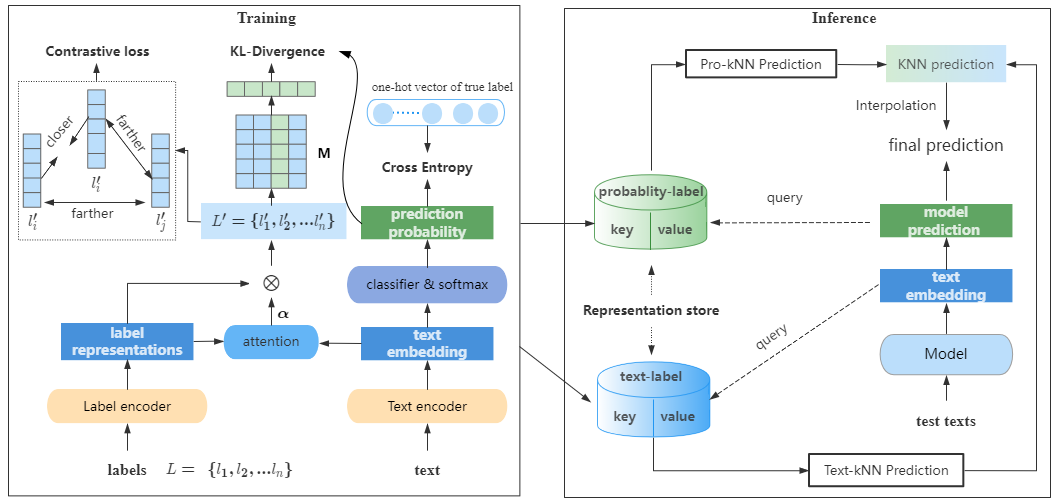}
\caption{The overall framework of our proposed method. The representation store contains a set of representation-label pairs, which are extracted from the hidden states of the label distribution learning enhanced model. During inference, when querying $k$ nearest neighbors from the representation store according to the similarity distance, the similarity distances are converted to $k$NN prediction distribution. Interpolating the $k$NN distribution with the vanilla model prediction distribution, we get the final distribution.}
\label{Figure2}
\vspace{-0.3cm}
\end{figure*}

\section{Method.}
\label{sec:method}
The overview of our proposed method is shown in Figure \ref{Figure2}, and this section describes our method in depth from its two primary parts. Firstly, we design a dual $k$ nearest neighbor (D$k$NN) framework including two $k$NN modules (text-$k$NN and pro-$k$NN) for the text classification model (single-label classification). Secondly, we propose a label distribution learning module based on label similarity to help the model distinguish texts more effectively. By training the model with this module, the D$k$NN can also be enhanced.

\subsection{Problem Formulation.}
Given an input text $\boldsymbol{x}=\left\{x_1,\dots,x_n\right\}$ with length $n$, $x_i (i=1,\dots,n)$ denotes the $i$-th word token within this text. $D =\left\{(\boldsymbol{x}_1,\boldsymbol{y}_1),\dots,(\boldsymbol{x}_N,\boldsymbol{y}_N)\right\}$ is training datasets with $N$ texts, 
where $(\boldsymbol{x_i,y_i})$ is a text sequence and corresponding one-hot label vector. For each text $\boldsymbol{x}$, most text classification models have a common paradigm to process it: use a deep neural network to obtain text embedding $\boldsymbol{h}$; then feed it to a simple linear layer with softmax activation function to obtain the predicted label distribution $\boldsymbol{p}$; finally calculate the cross entropy loss between  $\boldsymbol{p}$ and the true label vector $\boldsymbol{y}$. 

\subsection{$k$ Nearest Neighbor in Text Classification.}
To comprehensively obtain the information generated by each part of the models during the inference stage, we propose the D$k$NN framework, which includes the following steps: creating the key-value representation store, making two $k$NN module predictions based on it, and combining these two predicted distributions to get the final $k$NN prediction.

\subsubsection{Representation store.}
The representation store consists of a set of key-value pairs, and we construct two representation stores to explicitly memorize texts. To be specific, for each text-label pair 
$(\boldsymbol{x}_i,\boldsymbol{y}_i)$ from training datasets $D$,  
we extract the text embedding representation $\boldsymbol{h}_i$ and the predicted label 
probability representation $\boldsymbol{p}_i$ by a single forward pass over $\boldsymbol{x}_i$. Then the text representation store can be constructed as follows: 
$S_{tr} =\left\{\mathcal{K},\mathcal{V}\right\}=\left\{(\boldsymbol{h}_i,\boldsymbol{y}_i)\right\}_{i=1}^N$, and the predicted label probability representation store can be constructed as: 
$S_{pr} =\left\{\mathcal{K},\mathcal{V}\right\}=\left\{(\boldsymbol{p}_i,\boldsymbol{y}_i)\right\}_{i=1}^N$, where $\mathcal{K}$ is the key set and $\mathcal{V}$ is the corresponding value set. 

\subsubsection{Inference.} 
In the inference stage, given an input text $\boldsymbol{x}_i$, our model generates the predicted label probability distribution $\boldsymbol{p}_i$ and the text embedding $\boldsymbol{h}_i$. For $\boldsymbol{h}_i$, 
the text-$k$NN queries the text representation store $S_{tr}$ through the $L^2$ Euclidean distance $d_{L^2}(\cdot,\cdot)$ to obtain the $k$ nearest neighbors $\mathcal{N}_{tr}=\left\{({k}_i,{v}_i)\right\}_{i=1}^k \subseteq S_{tr}$. 
For $\boldsymbol{p}_i$, it is utilized by pro-$k$NN to query 
$k$ nearest neighbors $\mathcal{N}_{pr}=\left\{({k}_i,{v}_i)\right\}_{i=1}^k \subseteq S_{pr}$ from predicted label probability representation store $S_{pr}$ with Kullback–Leibler divergence (KL-divergence) $d_{KL}(\cdot,\cdot)$ \cite{kullback1951information} as distance measure. Then, these two $k$NN  predicted distributions over neighbors can be computed by applying a softmax to their negative distances and combining multiple occurrences of the same label item. 
\begin{equation}
\begin{aligned}
    \boldsymbol{p}_{text-knn}(\boldsymbol{y}_i|\boldsymbol{x}_i) \propto \sum_{(k_i,v_i) \in \mathcal{N}_{tr}} \mathbbm{1}_{y=v_i} {\rm exp}(-d_{L^2}(\boldsymbol{k_i},\boldsymbol{h_i}))
\end{aligned}
\end{equation}
\begin{equation}
\begin{aligned}
    \boldsymbol{p}_{pro-knn}(\boldsymbol{y}_i|\boldsymbol{x}_i) \propto \sum_{(k_i,v_i) \in \mathcal{N}_{pr}} \mathbbm{1}_{y=v_i} {\rm exp}(-d_{KL}(\boldsymbol{k_i},\boldsymbol{p_i}))
    \label{kl}
\end{aligned}
\end{equation}
Next, we improve high-confidence predictions while downgrading low-confidence ones by squaring and normalizing the two current predictions ($\boldsymbol{p}_{text-knn}$ and $\boldsymbol{p}_{pro-knn}$ can be viewed as $\boldsymbol{p}_{ij}$):
\begin{equation}
\begin{aligned}
    \boldsymbol{p}_{ij}^{\prime}= \frac{\boldsymbol{f}_j}{\sum_{j^{\prime}} \boldsymbol{f}_{j^{\prime}}} , \boldsymbol{f}_j = \frac{\boldsymbol{p}_{ij}^2}{\sum_{j} \boldsymbol{p}_{ij}}
\end{aligned}
\end{equation}
Finally, we interpolate the pure model prediction $\boldsymbol{p}_i$ and final $k$NN prediction $\boldsymbol{p}_{knn}(\boldsymbol{y}_i|\boldsymbol{x}_i)$ with a tuned parameter $\lambda$ to obtain final prediction results:
\begin{equation}
\begin{aligned}
    \boldsymbol{p}(\boldsymbol{y}_i|\boldsymbol{x}_i) = \lambda \boldsymbol{p}_{knn}(\boldsymbol{y}_i|\boldsymbol{x}_i)+(1-\lambda)\boldsymbol{p}_i
\end{aligned}
\end{equation}
\begin{equation}
\begin{aligned}
    \boldsymbol{p}_{knn}(\boldsymbol{y}_i|\boldsymbol{x}_i) = \frac{\boldsymbol{p}_{pro-knn}^{\prime}(\boldsymbol{y}_i|\boldsymbol{x}_i)+\boldsymbol{p}_{text-knn}^{\prime}(\boldsymbol{y}_i|\boldsymbol{x}_i)}{2}
\end{aligned}
\end{equation}
where $\boldsymbol{p}_{knn}$ is the average of the pro-$k$NN distribution $\boldsymbol{p}_{pro-knn}^{\prime}$ and the text-$k$NN distribution $\boldsymbol{p}_{text-knn}^{\prime}$.

\subsection{Label Distribution Learning.}
In our experiments, we find that the traditional $k$nn algorithm may cause prediction errors when the retrieved neighbors include noises, and cause over-fitting problems when the retrievals are most similar.
The underlying reason is that original text classification models can not effectively distinguish semantic differences between texts, particularly for those different texts with semantically similar labels, thus their text embeddings are less discriminative and have a negative impact on the $k$nn retrieval process. To address this problem, in the training stage, we propose a label distribution learning module based on label similarity to generate a distribution vector, which is a distinct representation with similarity simultaneously. In the following, we first define similarity measures among labels and then present how to equip the similarity measures to our training objective.

\subsubsection{Label Similarity.} 
Given texts and label names, we consider the similarities and differences semantically among various labels based on the texts. To utilize the semantic information of labels, we represent them by a label encoder, which is a deep neural network (DNN), to generate a trainable representation matrix 
$\boldsymbol{\mathrm{L}}=[\boldsymbol{l}_1,\boldsymbol{l}_2,\dots,\boldsymbol{l}_c]$ 
with length $c$ and dimension $d$, where $c$ is the number of label categories.
For an input text $\boldsymbol{x}_i$, and its text embedding $\boldsymbol{h}_i$ with dimension $d$, we measure the compatibility score for the entire label categories (i.e., the matching degree of texts and labels):
\begin{align}
    \boldsymbol{\alpha} = softmax(\boldsymbol{h}_i \boldsymbol{\mathrm{L}})
\end{align}
where $\bm \alpha \in \mathbb{R}^c$. The label representation can be updated by weighting text-based attention score:
\begin{align}
    \boldsymbol{\mathrm{L}}^{\prime}
           = [ \alpha_1 \boldsymbol{l_1},\alpha_2 \boldsymbol{l_2},\dots,\alpha_c \boldsymbol{l_c} ]
\end{align}
However, focusing solely on the attention between texts and labels ignores the relationship between inter-classes.

\subsubsection{Contrastive Learning with Label Similarity.} Our next goal is to encourage the model to be aware of the relationship between inter-classes, which can help the model to better represent the text with different labels. Typically, contrastive learning methods aim to teach models to distinguish between target samples and other samples. They have been proven to be beneficial for various tasks and have caught our attention. So, we first propose a straightforward and efficient method for constructing a label similarity matrix to represent the relationship between different labels:
\begin{align}
    \boldsymbol{\mathrm{M}} = \boldsymbol{\mathrm{L}}^{\prime} \boldsymbol{\mathrm{L}}^{\prime}{^{T}}
\end{align}
where $\boldsymbol{\mathrm{M}} =[\boldsymbol{m_1},\boldsymbol{m_2},\dots, \boldsymbol{m_c}]$ is the matrix of size $c \times c$,  each vector in $\boldsymbol{\mathrm{M}}$ represents the similarity value distribution between its corresponding label and all labels.
Then, we introduce a contrastive objective $\mathcal L_{\mathrm{CL}}$:
\begin{align}
    \mathcal L_{\mathrm{CL}} = \frac{\sum_{i=1}^{c}\sum_{j=1,j\neq i}^{c}max\left\{0,\rho-{\mathrm{M}}_{i,i}+{\mathrm{M}}_{i,j}\right\}}{c\times(c-1)}
    \label{contrasive learning} 
\end{align}
where $\rho \in [0,1]$ is a pre-defined margin, ${\mathrm{M}}_{i,j}$ is the value that denotes the similarity relationship between different labels and ${\mathrm{M}}_{i,i}$ is the self-similarity value. 
Intuitively, the model learns to pull away the distances between representations of distinct labels and distinguish texts better by training with $\mathcal L_{\mathrm{CL}}$.

\subsubsection{Training Objective.} 
For the ground-truth label, we first take out its corresponding column $\boldsymbol{m}$ from $\boldsymbol{\mathrm{M}}$ and transfer the similarity value distribution to similarity probability distribution by 
$\boldsymbol{q}= softmax(\boldsymbol{m})$, then we utilize KL-divergence object  
$\mathcal L_{\mathrm{KL}}$ to measure the distribution difference of $\boldsymbol{q}$ and $\boldsymbol{p}$. 
The $\mathcal L_{\mathrm{KL}}$ corresponds with Eq.\ref{kl}, which makes the model aware of the $k$nn retrieval process.
Denoting cross entropy loss as $\mathcal L_{\mathrm{CE}}$, the overall training objective is defined as:
\begin{align}
    \mathcal L = \mathcal L_{\mathrm{CE}} + \mathcal L_{\mathrm{KL}} + \mathcal L_{\mathrm{CL}}
\end{align}

\begin{figure}[h!]
    \begin{minipage}[t]{0.5\linewidth}
        \centering
        \includegraphics[width=\textwidth]{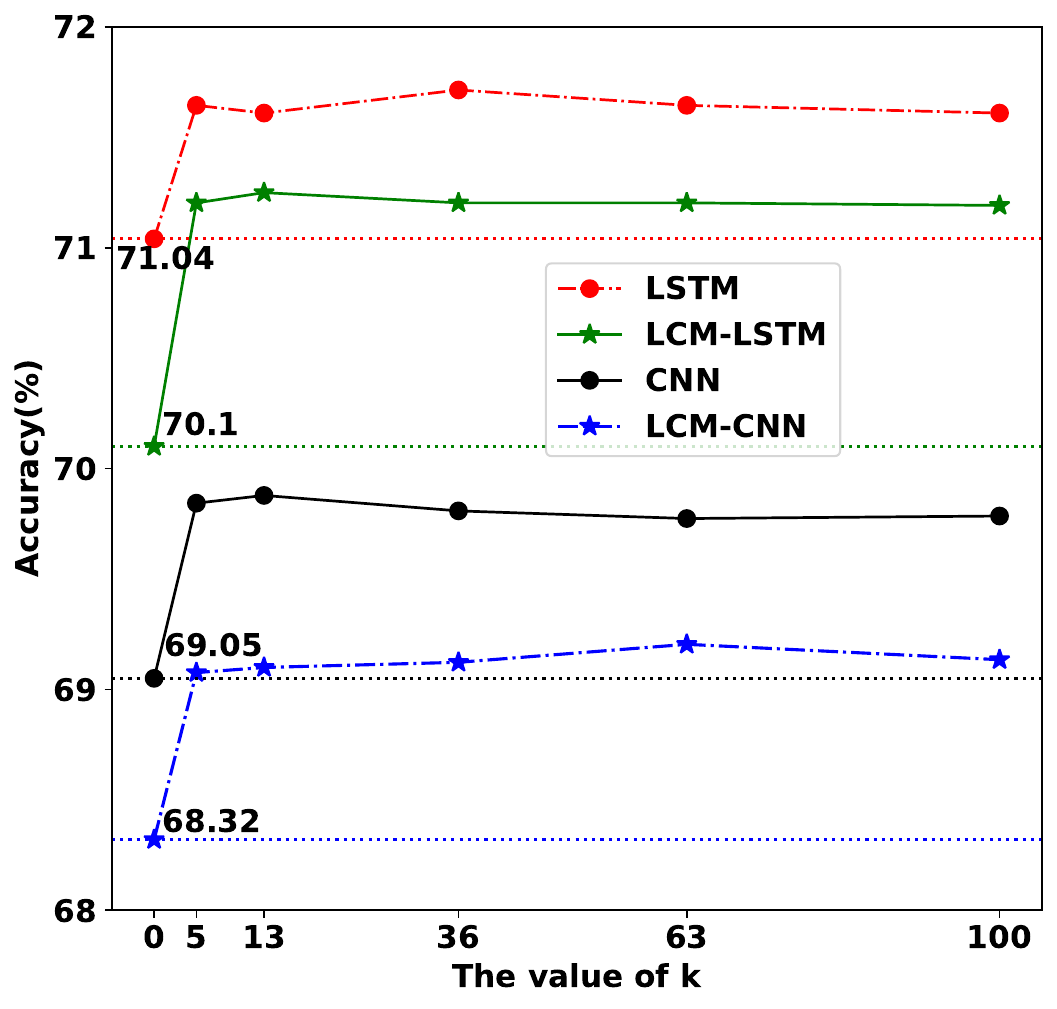}
        \centerline{(a) Analysis of the $k$}
    \end{minipage}%
    \begin{minipage}[t]{0.5\linewidth}
        \centering
        \includegraphics[width=\textwidth]{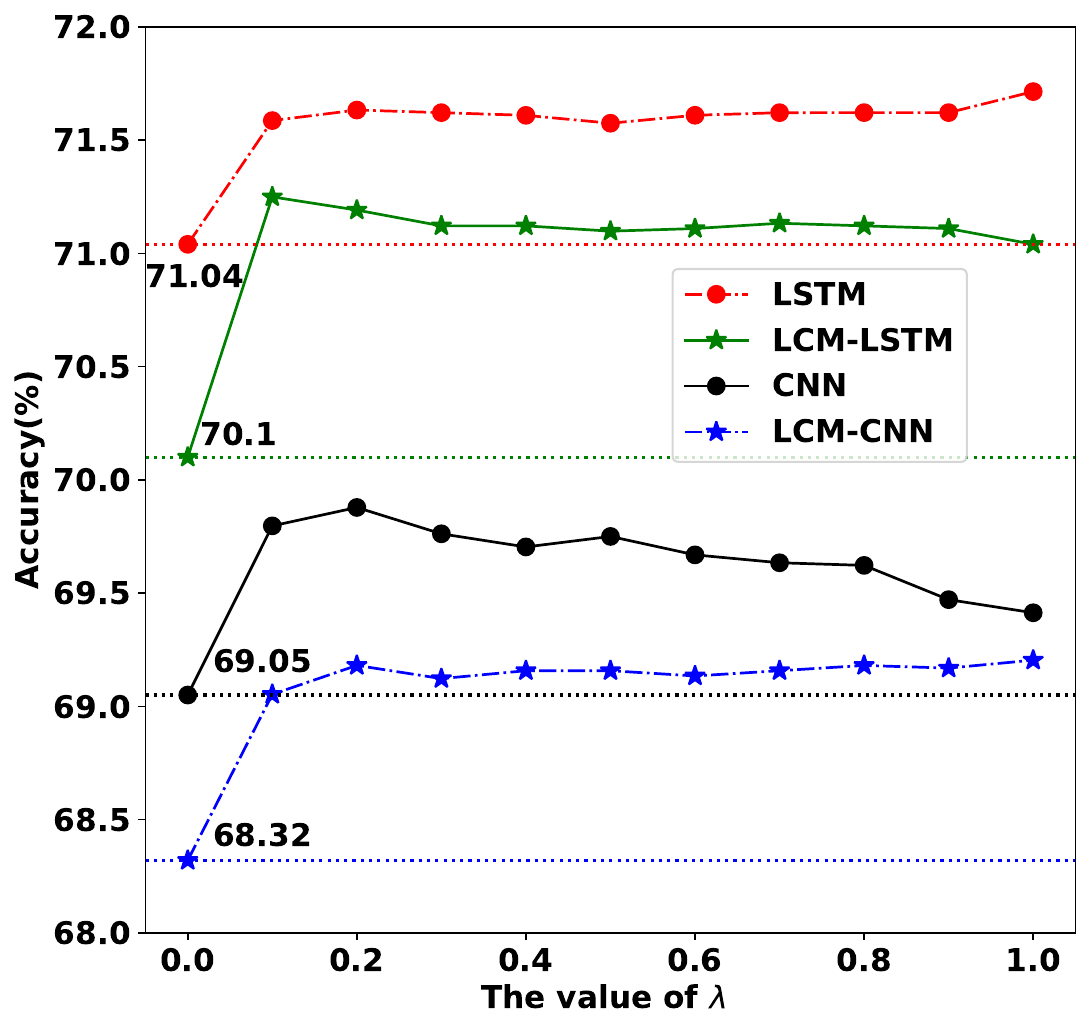}
        \centerline{(b) Analysis of the $\lambda$}
    \end{minipage}
    \caption{Hyperparameter settings of the D$k$NN.}
    \label{Figure 2}
\vspace{-0.5cm}
\end{figure}

\begin{table*}
\footnotesize
\centering
\setlength\tabcolsep{13pt}
\renewcommand\arraystretch{0.9} 
\begin{tabular}{l c c c c c}
\toprule[1.3pt]
\rule{0pt}{9pt}
\multirow{2}{*}{\textbf{Models}} &
\multicolumn{5}{c}{\makecell[c]{\textbf{Accu. (\%)}}} \\
\cline{2-6}
\rule{0pt}{13pt}
&\makecell{\textbf{WOS-5736} } & \makecell{\textbf{20NG}}  & \makecell{\textbf{AGNews}} & \makecell{\textbf{DBPedia} }& \makecell{\textbf{FDCNews}} \\
\toprule[1.3pt]
CNN \cite{DBLP:conf/emnlp/Kim14} & 65.15 $_{\pm 0.0158}$ & 72.22 $_{\pm 0.0046}$  & 89.73 $_{\pm 0.0033}$ & 96.09 $_{\pm 0.0011}$ & 80.78 $_{\pm 0.0096}$ \\
Ours & \textbf{69.88 $_{\pm 0.0164}$} & \textbf{77.12 $_{\pm 0.0034}$} & \textbf{90.33 $_{\pm 0.0013}$} & \textbf{96.92 $_{\pm 0.0009}$} & \textbf{82.07 $_{\pm 0.0056}$} \\
\midrule[1.0pt]
LCM-CNN \cite{DBLP:conf/aaai/GuoH0HL21} & 67.20 $_{\pm 0.0037}$ & 76.44 $_{\pm 0.0075}$ & 89.99 $_{\pm 0.0023}$ & 96.96 $_{\pm 0.0005}$ & 82.17 $_{\pm 0.0039}$\\
Ours & \textbf{69.20 $_{\pm 0.0117}$} & \textbf{77.20 $_{\pm 0.0046}$} & \textbf{90.33 $_{\pm 0.0014}$} & \textbf{97.11 $_{\pm 0.0011}$} & \textbf{82.39 $_{\pm 0.0020}$}\\ 
\midrule[1.0pt]
LSTM \cite{Liu_Qiu_Huang_2016} &  69.29 $_{ \pm 0.0160}$ & 69.84 $_{\pm 0.0133}$ & 88.52 $_{\pm 0.0040}$ & 93.44 $_{\pm 0.0171}$ & 77.59 $_{\pm 0.0073}$ \\
Ours & \textbf{71.71 $_{\pm 0.0061}$} & \textbf{75.09 $_{\pm 0.0102}$} & \textbf{89.27 $_{\pm 0.0023}$} & \textbf{94.91 $_{\pm 0.0018}$} & \textbf{80.44 $_{\pm 0.0040}$} \\
\midrule[1.0pt]
LCM-LSTM \cite{DBLP:conf/aaai/GuoH0HL21} & 69.32 $_{\pm 0.0157}$ & 74.00 $_{\pm 0.0046}$ & 88.12 $_{\pm 0.0049}$ & 94.09 $_{\pm 0.0124}$ & 79.78 $_{\pm 0.0096}$\\
Ours & \textbf{71.25 $_{\pm 0.0107}$} & \textbf{74.77 $_{\pm 0.0102}$} &  \textbf{89.10 $_{\pm 0.0069}$} & \textbf{94.34 $_{\pm 0.0098}$} & \textbf{80.64 $_{\pm 0.0016}$} \\
\midrule[1.0pt]
LCM-BERT \cite{DBLP:conf/aaai/GuoH0HL21} & 78.33 $_{\pm 0.0155}$ & 90.03 $_{\pm 0.0031}$
& 91.30 $_{\pm 0.0012}$ & 97.95 $_{\pm 0.0009}$ & 96.15 $_{\pm 0.0040}$\\
Ours & \textbf{79.05} $_{\pm 0.0142}$ & \textbf{90.38 $_{\pm 0.0030}$} & \textbf{91.42 $_{\pm 0.0004}$} & \textbf{98.32 $_{\pm 0.0008}$} & \textbf{96.39 $_{\pm 0.0064}$}\\
\midrule[1.0pt]
ALBERT \cite{DBLP:conf/iclr/LanCGGSS20} & 76.49 $_{\pm 0.0223}$ & 88.27 $_{\pm 0.0129}$ & 89.71 $_{\pm 0.0014}$ & 96.74 $_{\pm 0.0080}$ &  94.30 $_{\pm 0.0029}$ \\
Ours & \textbf{77.27 $_{\pm 0.0110}$} &  \textbf{89.34 $_{\pm 0.0053}$} &  \textbf{92.04 $_{\pm 0.0019}$} & \textbf{98.16 $_{\pm 0.0026}$} &  \textbf{96.15 $_{\pm 0.0066}$} \\
\bottomrule[1.0pt]
BERT \cite{DBLP:journals/corr/abs-2110-02523} & 77.79 $_{\pm 0.0117}$ & 88.21 $_{\pm 0.0005}$ & 90.24 $_{\pm 0.0014}$ & 97.49 $_{\pm 0.0011}$ & 96.63 $_{\pm 0.0016}$ \\
Ours & \textbf{78.95 $_{\pm 0.0113}$} & \textbf{90.53 $_{\pm 0.0016}$} & \textbf{91.35 $_{\pm 0.0014}$} & \textbf{98.31 $_{\pm 0.0009}$} & \textbf{97.59 $_{\pm 0.0028}$}\\
\midrule[1.0pt]

\multicolumn{6}{c}{\makecell[c]{\textbf{Other baselines based on KNN or contrastive learning}}} \\
\midrule[1.0pt]

Base & 77.79 $_{\pm 0.0117}$ & 88.21 $_{\pm 0.0005}$ & 90.24 $_{\pm 0.0014}$ & 97.49 $_{\pm 0.0011}$ & 96.63 $_{\pm 0.0016}$ \\
SCL \cite{Gunel_Du_Conneau_Stoyanov_2021} & 77.90 $_{\pm 0.6077}$ & 88.45 $_{\pm 0.2592}$ & 90.28 $_{\pm 0.0809}$ & 97.58 $_{\pm 0.0722}$ & 96.65 $_{\pm 0.1841}$ \\ 
SCL-MoCo \cite{He_Fan_Wu_Xie_Girshick_2020} & 78.01 $_{\pm 0.3389}$ & 88.46 $_{\pm 0.1750}$ & 90.32 $_{\pm 0.1153}$ & 97.64 $_{\pm 0.0861}$ & 96.97 $_{\pm 0.2922}$\\
KNN-BERT \cite{DBLP:journals/corr/abs-2110-02523} & 78.11 $_{\pm 0.0093}$ & 88.97 $_{\pm 0.0035}$ & 90.38 $_{\pm 0.0013}$ & 97.71 $_{\pm 0.0014}$ & 97.42 $_{\pm 0.0022}$ \\
Ours & \textbf{78.95 $_{\pm 0.0113}$} & \textbf{90.53 $_{\pm 0.0016}$} & \textbf{91.35 $_{\pm 0.0014}$} & \textbf{98.31 $_{\pm 0.0009}$} & \textbf{97.59 $_{\pm 0.0028}$}\\
\bottomrule[1.3pt]
\end{tabular}
\caption{Test accuracy on different datasets. We run all models and report their mean$\pm$ standard deviation. The results with outstanding improvement over the base model are bolded, and these values indicate statistically significantly better (based on the student $t$-test, p $<$ 0.05) performances across the board.}
\label{Table 1}
\vspace{-0.3cm}
\end{table*}

\section{Experiment.}
\subsection{Baselines.}
We first adopt the following models as our baselines and apply our method to all of them: CNN \cite{DBLP:conf/emnlp/Kim14}, LSTM \cite{Liu_Qiu_Huang_2016}, BERT \cite{DBLP:conf/naacl/DevlinCLT19}, ALBERT \cite{DBLP:conf/iclr/LanCGGSS20}, and LCM \cite{DBLP:conf/aaai/GuoH0HL21}. LCM is an enhancement component to current popular text classification models, which is similar to our proposed label distribution learning module. 
For LCM, we follow \cite{DBLP:conf/aaai/GuoH0HL21} to implement it based on LSTM, CNN, and BERT. Then, we compare our $k$NN-based approach with KNN-BERT \cite{DBLP:journals/corr/abs-2110-02523}. Specifically, KNN-BERT utilizes the traditional $k$NN classifier in pretrained model fine-tuning and trains clustered representations based on a supervised contrastive learning framework to improve the performance of pretrained model fine-tuning. Further, we follow KNN-BERT to select a few related works SCL \cite{Gunel_Du_Conneau_Stoyanov_2021} and SCL-MoCo \cite{He_Fan_Wu_Xie_Girshick_2020} as other baselines, and use the same backbone (BERT) for these baseline methods. For a fair comparison, we train them using the same parameters with ours.
 

\subsection{Datasets and Evaluation Metric.}
We evaluate the effectiveness and robustness of our proposed model on the following several text classification datasets: 20NG dataset\footnote{\url{http://qwone.com/~jason/20Newsgroups/}} \cite{DBLP:conf/icml/Lang95} is an English news dataset categorized into 20 different categories, in total, it contains 18846 documents; AGNews dataset\footnote{\url{http://groups.di.unipi.it/ gulli/AG corpus of news articles.html}} \cite{DBLP:conf/nips/ZhangZL15} consists of news articles, which contains 127600 documents pertaining to the four largest classes;
DBPedia\footnote{\url{http://dbpedia.org}} \cite{DBLP:conf/nips/ZhangZL15} is an ontology classification dataset containing 630000 documents over 14 categories;
FDCNews dataset\footnote{\url{http://www.nlpir.org}} is a Chinese dataset constructed by Fudan University, which has 9833 documents divided into 20 categories; 
WOS-5736\footnote{\url{http://archive.ics.uci.edu/index.php}} \cite{DBLP:conf/icmla/KowsariBHMGB17} is an academic paper dataset, which includes 5736 documents classified into 3 coarse-grained categories and 11 fine-grained categories. We conduct experiments using fine-grained categories. We follow the data processing of \cite{DBLP:conf/aaai/GuoH0HL21}, then train the model with a train set and evaluate on development set after every epoch.

\subsection{Parameter Settings.}
In our experiments, we set the batch size to 128. For non-pretrained models, we set the maximum sequence lengths as 100 tokens and the embedding size as 64. For CNN, we use two convolution blocks, the number of filters for each block is 64 and 32 respectively, and the filter sizes are both 3. For BERT and ALBERT, the maximum sequence length is 128 tokens and the hidden size is 768. For LCM, we use the default parameter settings in the original papers. All models are trained with the Adam Optimizer \cite{DBLP:journals/corr/KingmaB14} on GTX 2080 Ti and RTX 3090.

\subsection{Experiment Results.}
In our experiments, we split each dataset into different train and test sets 5 times with a 7:3 splitting ratio. Then we evaluate all models 5 times on these various train test splitting and report the average performance.  We use accuracy as an evaluation metric following most previous works.

\subsubsection{Overall Performance.} The results of our method compared to other methods are listed in Table \ref{Table 1}. The LCM-X means that the LCM model uses X as a basic predictor. As shown in Table \ref{Table 1}, our method can improve the performance of baseline models on all datasets. For instance,  
our method improves the accuracy of CNN by 4.73$\%$ on WOS-5736 and 4.9$\%$ on 20NG. The performance improvements on LSTM are also quite obvious with 5.25$\%$ on 20NG and 2.85$\%$ on FDCNews. For pre-trained models, we also further compare our method with other baselines based on $k$NN and contrastive learning. As seen, our method performs better than them, which further verifies the effectiveness and applicability of our proposed method.

\subsubsection{Ablation Test.}
As mentioned above, our method mainly includes the following parts: a label distribution learning module (denoted as LL) and a dual $k$ nearest neighbor framework (denoted as D$k$NN). We perform an ablation test to demonstrate the effectiveness of each part. As shown in Table 
\ref{tab-aba}, D$k$NN and LL can both generally enhance the performance of the baseline models. Even though the retrieval effect of D$k$NN remains consistent on some baseline models, when combined with our label distribution learning module, the improvements brought by the D$k$NN have significantly improved. This demonstrates that the D$k$NN is successfully enhanced by our LL.

\section{Discussion.}
\subsection{Analysis for D$k$NN.}
The $k$NN algorithm in the existing work makes predictions based on cached text embeddings and corresponding target tokens. 
We emphasize the importance of calculating and comparing the predicted label probability distribution distance between the samples to be classified and each historical sample, thus we introduce D$k$NN consisting of text-$k$NN (denoted as t-$k$NN) and pro-$k$NN (denoted as p-$k$NN). Here, we conduct a parameter analysis of D$k$NN on the WOS-5736 dataset, and an ablation study to explore the effectiveness of D$k$NN.

\subsubsection{Hyperparameters in D$k$NN } 
In our proposed D$k$NN, $k$ is a significant hyperparameter that returns top-$k$ nearest neighbors for each query. As shown in Figure \ref{Figure 2}(a), the hyperperformance with D$k$NN is always better than that when using only the model output ($k$= 0), which further verifies the effectiveness of D$k$NN to utilize the information from the most similar historical samples.
Additionally, the performance initially increases quickly, but as the $k$ increases, it tends to plateau or even decline. That can be summed up that a large $k$ is not necessary and a small $k$ can still have a considerable but not infallible result, which can save the computational cost of retrieval. Moreover, we use a hyperparameter $\lambda$ to interpolate between 
the distribution from D$k$NN search over the dataset and the pure model prediction distribution during inference. Figure \ref{Figure 2}(b) demonstrates the trend of model performance with $\lambda$. The general trend is similar to that of $k$, which further demonstrates the performance is poor when relying simply on the model prediction ($\lambda $= 0). We also notice that LSTM and LCM-CNN can obtain the greatest results when only using the model predictions ($\lambda$=1). 

\begin{table}[ht]
\footnotesize
\centering
\setlength\tabcolsep{11pt}
\begin{tabular}{l| c c}
\toprule[1.3pt]
\makecell[l]{\textbf{Models}} & \makecell[c]{\textbf{WOS-5736}} & \makecell[c]{\textbf{20NG}}\\ 

\midrule
CNN & 65.15 & 72.22\\
CNN+D$k$NN & 65.24 & 72.54\\
CNN+LL & 68.54 & 76.59\\
CNN+LL+D$k$NN & \textbf{69.88} & \textbf{77.12}\\

\midrule
LCM-CNN & 67.20 & 76.44\\
LCM-CNN+D$k$NN & 67.45 & 76.94\\
LCM-CNN+LL & 68.32 & 76.57\\
LCM-CNN+LL+D$k$NN & \textbf{69.20} & \textbf{77.20}\\

\midrule
LSTM & 69.29 & 69.84\\
LSTM+D$k$NN & 69.37 & 70.01\\
LSTM+LL & 70.45 & 74.12\\
LSTM+LL+D$k$NN & \textbf{71.71} & \textbf{75.09}\\
\midrule
LCM-LSTM &  69.32 & 74.00\\
LCM-LSTM+D$k$NN & 69.45 & 74.01\\
LCM-LSTM+LL & 70.10 & 74.16\\
LCM-LSTM+LL+D$k$NN & \textbf{71.25} & \textbf{74.77}\\

\midrule
BERT & 77.79 & 88.21\\
BERT+D$k$NN & 78.13 & 88.98\\ 
BERT+LL & 78.45 & 90.13\\ 
BERT+LL+D$k$NN & \textbf{78.95} & \textbf{90.53}\\
\midrule
LCM-BERT & 78.33  & 90.03\\
LCM-BERT+D$k$NN & 78.37 & 90.05\\
LCM-BERT+LL & 78.93 & 90.23\\
LCM-BERT+LL+D$k$NN & \textbf{79.05} & \textbf{90.38}\\
\bottomrule[1.3pt]
\end{tabular}
\caption{Accuracy ($\%$) of the ablation tests for main components in our method. D$k$NN means the dual $k$ nearest neighbor framework proposed in our method, and LL denotes the label distribution learning module.}
\label{tab-aba}
\vspace{-0.6cm}
\end{table}

\begin{table}[htb!]
\setlength\tabcolsep{6pt}
\footnotesize
\centering
\begin{tabular}{l| c c c}
\toprule[1.3pt]
\makecell[l]{\textbf{Models}} & \makecell[c]{\textbf{D$k$NN}} 
& \makecell[c]{\textbf{w/o p-$k$NN}} & \makecell[c]{\textbf{w/o t-$k$NN}}\\ 
\midrule
LSTM  & \textbf{69.37} & 69.27 & 69.04 \\ 
LSTM+LL  & \textbf{71.71} & 71.66 & 71.28\\ 
\midrule
LCM-LSTM & \textbf{69.45} & 69.41 & 68.90 \\
LCM-LSTM+LL & \textbf{71.25} & 71.14 & 71.01  \\
\midrule
CNN & \textbf{69.37} & 69.27 & 69.04\\
CNN+LL & \textbf{69.88} & 69.03 & 69.06\\
\midrule
LCM-CNN & \textbf{67.45} & 67.37 & 67.40\\
LCM-CNN+LL & \textbf{69.20} & 69.06 & 68.93\\
\midrule
BERT & \textbf{78.13} & 78.00 & 78.06 \\
BERT+LL & \textbf{78.95} & 78.58 & 78.66 \\
\midrule
LCM-BERT & \textbf{78.37} & 78.16 & 78.08 \\
LCM-BERT+LL & \textbf{79.05} & 78.92 & 78.93 \\
\bottomrule[1.3pt]
\end{tabular}

\caption{Results of accuracy (\%) on the WOS-5736 datasets. We experiment ablation tests for D$k$NN.}
\label{table3}
\vspace{-0.6cm}
\end{table}

\subsubsection{Effectiveness of D$k$NN.} 
In our paper, we use text-$k$NN to measure text embedding similarities and pro-$k$NN to calculate the distance between probability distributions. To verify the effectiveness of these two modules, we further perform ablation experiments on the WOS-5736 dataset. From Table \ref{table3}, we observe that
pro-$k$NN (denoted as p-$k$NN) is as important as text-$k$NN (denoted as t-$k$NN) to obtain a better performance. After removing p-$k$NN or t-$k$NN, the performances will drop to a large degree. In some experiments, the improvements brought by p-$k$NN outperform t-$k$NN, 
which further proves that retrieving neighbors 
according to the prediction probability is also an 
appropriate way.

\begin{figure*}[htp!]
\centering
\includegraphics[width=17.5cm,height=3.5cm]{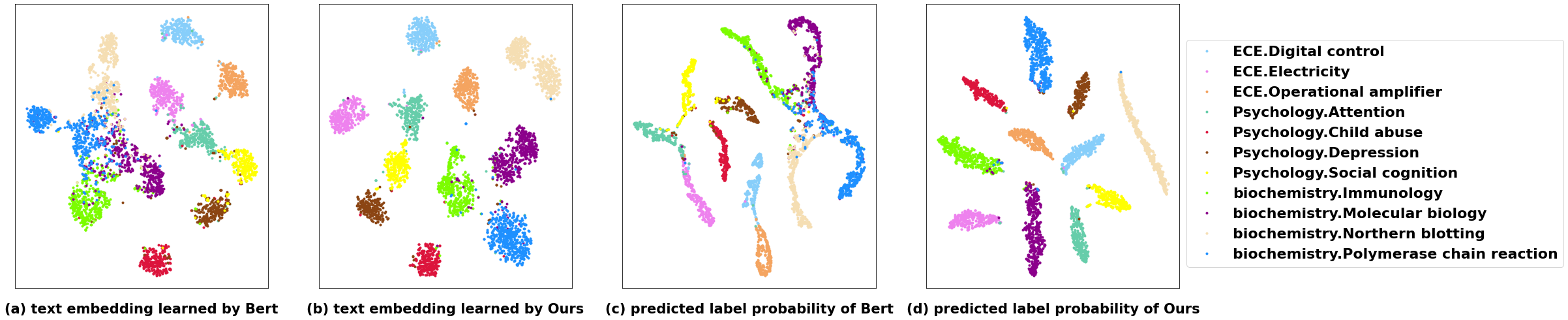}
\caption{visualize the cached in the representation store by t-SNE tool.}
\vspace{-0.2cm}
\label{fig:t-SNE}
\end{figure*} 

\begin{table*}[t]
\footnotesize
\setlength\tabcolsep{14pt}
\centering
\begin{tabular}{l|c|c|c}
\toprule[1.3pt]
\makecell[l]{} & \makecell[c]{\textbf{biochemistry}}& \makecell[c]{\textbf{ECE}} & \makecell[c]{\textbf{Psychology}} \\ 
\midrule
\textbf{Labels} & \makecell[l]{biochemistry.Polymerase chain reaction \\ biochemistry.Molecular biology \\ biochemistry.Northern blotting \\ biochemistry.Immunology} & \makecell[l]{ECE.Electricity \\ ECE.Digital control \\ ECE.Operational amplifier} & 
\makecell[l]{Psychology.Attention \\ Psychology.Child abuse \\ Psychology.Social cognition \\ Psychology.Depression} \\
\bottomrule
\end{tabular}
\caption{Labels of the WOS-5736 dataset}
\label{tab-lables}
\vspace{-0.2cm}
\end{table*}

\subsection{Analysis for LL.}
In our method, the LL part generates the learned label distribution, which explores the similarity relationship among labels by taking into account the complicated relationship between the input texts and labels. Then we introduce this label distribution representation vector into the calculation of our training objective. There are two reasons for this operation: (1) The $k$ Nearest Neighbor mechanism relies heavily on the training sample set and has poor fault tolerance to the training data. If some samples in the training datasets are mislabeled and just right next to the text to be classified, they will directly lead to the inaccuracy of the predicted result. In this case, solely using cross entropy with one-hot label representation during training will make models more susceptible to the effects of noisy data (mislabeled data). However, the distribution generated by LL is based on similarity among different labels, by measuring the distance between this distribution and the predicted label probability distribution with KL-divergence loss function (denoted as $\mathcal L_{\rm KL}$), the value on the index of the wrong label will be alleviated. (2) During the inference stage, we also use KL-divergence as the distance measure to query $k$ nearest neighbors from the predicted label probability representation store, which exactly corresponds to the $\mathcal L_{\rm KL}$ in the overall training objective. Intuitively, by using LL, the predicted label probability cached in the representation store will be better, and the quality of the retrieved neighbors will also be improved.

\subsubsection{Impact of Training Objective.} 
We further conduct experiments to analyze the impact of our training objective. As shown in Table \ref{table6}, we equip $\mathcal L_{\rm KL}$ and contrastive loss objective $\mathcal L_{\rm CL}$
separately with the original cross entropy loss function (denoted as $\mathcal L_{\rm CE}$). From the results, we can see both $\mathcal L_{\rm KL}$ objective and $\mathcal L_{\rm CL}$ objective can effectively improve the performance.

\begin{table}[ht]
\footnotesize
\setlength\tabcolsep{6pt}
\centering
\begin{tabular}{l| c c c c c}
\toprule[1.3pt]

\makecell[l]{\textbf{Models}} & \makecell[c]{\textbf{0\%}}& \makecell[c]{\textbf{3\%}} & \makecell[c]{\textbf{6\%}} & \makecell[c]{\textbf{30\%}} & \makecell[c]{\textbf{50\%}}\\ 
\midrule
Bert &  77.79 & 77.11 & 75.78 & 69.74 & 57.86\\
+D$k$NN & \textbf{78.13} & \textbf{77.26} & \textbf{75.99} & \textbf{70.35} & \textbf{59.00} \\
\midrule
Increase & 0.34 & 0.15 & 0.21 & 0.61 & 1.14  \\
\midrule
Bert+LL & 78.45 & 78.16 & 77.27 & 69.76 & 58.36\\
+D$k$NN & \textbf{78.95} & \textbf{78.42} &  \textbf{77.57} & \textbf{70.83} & \textbf{59.50} \\
\midrule
Increase & 0.5 & 0.26 & 0.3 & 1.07 & 1.26  \\
\midrule
LCM-Bert & 78.33 & 77.44 & 76.27 & 70.85 & 59.85\\
+D$k$NN & \textbf{78.37} & \textbf{77.56} & \textbf{76.47} & \textbf{71.11} & \textbf{61.08}\\
\midrule
Increase & 0.04 & 0.12 & 0.2 & 0.26  &  1.23 \\
\midrule
LCM-Bert+LL & 78.93 & 78.37 & 77.26 &  69.94 & 58.06\\
+D$k$NN  & \textbf{79.05} & \textbf{78.55} & \textbf{77.56} & \textbf{70.75} & \textbf{59.31}\\
\midrule
Increase & 0.12 & 0.18 & 0.3 & 0.81 & 1.25  \\
\bottomrule[1.3pt]

\end{tabular}

\caption{The experimental results (accuracy, $\%$) on the noisy datasets, where the percentage (0\%, 3\%, 6\%, 30\%, 50\%) represents the proportion (noisy ratio) of randomly mislabeled samples. 
}
\label{table5}
\vspace{-0.2cm}
\end{table}

\begin{table}[htb!]
\centering
\footnotesize
\setlength\tabcolsep{16pt}
\begin{tabular}{l| c c}
\toprule[1.3pt]

\makecell[l]{\textbf{Bert}} & \makecell[c]{\textbf{WOS-5736}} & \makecell[c]{\textbf{20NG}}\\ 
\midrule
$\mathcal L_{\rm CE}$ & 77.79  & 89.51\\
$\mathcal L_{\rm CE}+\mathcal L_{\rm KL}$ & 78.23  & 90.10\\
$\mathcal L_{\rm CE}+\mathcal L_{\rm CL}$ & 78.13  & 89.86\\
$\mathcal L_{\rm CE}+\mathcal L_{\rm CL}+\mathcal L_{\rm KL}$ & \textbf{78.45} & \textbf{90.13}\\
\bottomrule[1.3pt]
\end{tabular}

\caption{Ablation over the KL-divergence object and contrastive learning objective of the proposed LL module using the Bert model on WOS-5736 dataset and 20NG datasets.}
\label{table6}
\vspace{-0.3cm}
\end{table}

\subsubsection{Representation Store Visualizations.} 
To explore whether the text embeddings and the predicted label probability are actually better by our LL, we use the t-SNE tool \cite{van2008visualizing} to visualize these representations cached in the representation store. Figure \ref{fig:t-SNE}a and Figure \ref{fig:t-SNE}b respectively show the visualization of text embeddings learned by the original Bert model and LL enhanced Bert model on the WOS-5735 datasets. We also plot the predicted label probability of the original Bert model and our model in Figure \ref{fig:t-SNE}c and Figure \ref{fig:t-SNE}d. As shown in Figure \ref{fig:t-SNE}a and Figure \ref{fig:t-SNE}c, we can find that the text embeddings and predicted label probability produced by the original model can be easily confused, especially those with similar labels. After using LL, the distance between different categories of texts becomes larger, which is shown in Figure \ref{fig:t-SNE}b and Figure \ref{fig:t-SNE}d. Even though some samples are still jumbled in the incorrect cluster, the boundary is more distinct and separate from the other categories. In this case, the neighbors retrieved by D$k$NN during inference are more accurate. 
In Figure \ref{fig:t-SNE}, we also notice that the representation of text embeddings is more horizontal and the predicted label probability representation is more vertical, which shows our D$k$NN can consider the distance among representations from more directions when retrieving in representation store.

\subsection{Performance on Noisy Datasets.}
Current text classification models require a large number of human-labeled texts as training data. Label noise is unavoidable when collecting data. As we all know, noisy data will severely reduce classification performance. Besides, the $k$NN mechanism is also sensitive to noisy data that has labeling errors. In the WOS-5735 dataset, we found that there are 11 fine-grained labels but they belong to 3 coarse-grained categories. As shown in Table \ref{tab-lables}, the labels that belong to the same category are similar in some way and hard to distinguish, which makes labeling errors more likely to occur and makes models overfit. To verify the robustness of our method against noise attacks, we generated some noisy datasets from the WOS-5735 dataset with different percentages of noise, and the mislabeling samples are all drawn from the same category to make the noisy datasets closer to reality. Then we conduct experiments with Bert as the base model. The results are shown in Table \ref{table5}, we can see that our LL module still enhances the $k$NN mechanism on the datasets with obvious label errors, and the increase of our method outperforms the improvement of the base model and LCM to a large degree. Even without D$k$NN, the LL module in our method still performs significantly better than LCM on noisy datasets. Meanwhile, this observation also proves that the proposed LL can make the model robust enough.

\section{Conclusion.}
In this paper, we propose a dual $k$NN framework (D$k$NN) with two $k$NN modules for current text classification models. 
 Then we introduce a label distribution learning module based on label similarity with contrastive learning to improve the performance of original models directly, as well as indirectly improve D$k$NN retrieval quality. 
 By conducting extensive experiments, we confirm the effectiveness of our method and analyze the performance improvement brought about by each module. 
 Besides, our method has higher robustness in noisy environments. In the future, we would like to explore two avenues. Firstly, designing a better framework to improve retrieval efficiency and reduce the size of the representation store for D$k$NN. Secondly, we can generalize our method to other classification tasks. 

\section*{Acknowledgments}
This work is supported by the China Knowledge Centre for Engineering Sciences and Technology (CKCEST-2022-1-7).
\bibliographystyle{elsarticle-num}

\bibliography{byuan}

\end{document}